\pdfoutput=1
\documentclass[11pt]{article}

\usepackage[]{EMNLP2023}

\usepackage{times}
\usepackage{latexsym}

\usepackage[T1]{fontenc}
\usepackage[utf8]{inputenc}

\usepackage{microtype}
\usepackage{amsmath}
\usepackage{graphicx}
\usepackage{multirow}
\usepackage{soul}

\usepackage{pifont}

\usepackage{times}
\usepackage{latexsym}
\usepackage{graphicx}
\usepackage{graphics}
\usepackage{booktabs}
\usepackage{subcaption}
\usepackage[ruled,vlined]{algorithm2e}
\usepackage{multirow}
\usepackage{tabularx}
\usepackage{amsmath}
\usepackage{bbold}
\usepackage{bm}
\usepackage{mathtools}
\usepackage{xcolor}
\usepackage{booktabs}
\usepackage{tabularx,ragged2e}
\usepackage{enumitem}
\usepackage{amssymb}
\usepackage{color, colortbl}
	
\definecolor{LightCyan}{rgb}{0.88,1,1}

\newcommand*\samethanks[1][\value{footnote}]{\footnotemark[#1]}

\usepackage{xcolor}
\usepackage{inconsolata}

\title{Can NLP Models `Identify', `Distinguish', and `Justify' Questions that Don't have a Definitive Answer?}

\author{Ayushi Agarwal\thanks{~~Equal Contribution, Contact: nppatel7@asu.edu} \hspace{9pt} ~~ 
Nisarg Patel\samethanks \hspace{15pt} ~~ 
Neeraj Varshney\samethanks \hspace{15pt} ~~ 
Mihir Parmar ~~ \\
\textbf{Pavan Mallina}\hspace{15pt} ~~ 
\textbf{Aryan Shah} \hspace{15pt}~~ 
\textbf{Srihari Raju Sangaraju} ~~ \\
\textbf{Tirth Patel} \hspace{15pt}~~ 
\textbf{Nihar Thakkar} \hspace{15pt}~~
\textbf{Chitta Baral}
  \\ \\
  Arizona State University \\
  }
\begin{document}
\maketitle
\begin{abstract}
Though state-of-the-art (SOTA) NLP systems have achieved remarkable performance on a variety of language understanding tasks, they primarily focus on questions that have a correct and a definitive answer. However, in real-world applications, users often ask questions that don't have a definitive answer.
Incorrectly answering such questions certainly hampers a system's reliability and trustworthiness. 
Can SOTA models accurately identify such questions and provide a reasonable response?

To investigate the above question, we introduce \texttt{QnotA}, a dataset consisting of five different categories of questions that don't have definitive answers.
Furthermore, for each \texttt{QnotA} instance, we also provide a corresponding \texttt{QA} instance i.e. an alternate question that ``\textit{can be}'' answered.
With this data, we formulate three evaluation tasks that test a system's ability to `\textit{identify}', `\textit{distinguish}', and `\textit{justify}' \texttt{QnotA} questions.
Through comprehensive experiments, we show that even SOTA models including GPT-3 and Flan T5 do not fare well on these tasks and lack considerably behind the human performance baseline.
We conduct a thorough analysis which further leads to several interesting findings.
Overall, we believe our work and findings will encourage and facilitate further research in this important area and help develop more robust models.

\end{abstract}

\section{Introduction}
%motivation

% \begin{figure}[!h]  
%     \centering
%     \includegraphics[height=8cm,width=7.44cm]{Pictures/T3.png}
%     \caption{Caption} 
%     \label{fig:teaser}
% \end{figure}

% \begin{figure}[!h]  
%     \centering
%     \includegraphics[height=6.4cm,width=8cm]{Pictures/teaser_temp.pdf}
%     \caption{Caption} 
%     \label{fig:teaser}
% \end{figure}

% \Neeraj{include papers of different reasoning skills}

% Natural Language Processing (NLP) models have achieved impressive results in various language tasks, including question answering. 
Recent advancements in Natural Language Processing (NLP) have led to the development of Question-Answering systems that possess remarkable capabilities of providing fluent and comprehensive answers to a wide range of questions \cite{khashabi-etal-2020-unifiedqa,NEURIPS2020_1457c0d6,zhang2022opt,lourie2021unicorn,chowdhery2022palm,rae2021scaling}. 
However, these systems primarily focus on questions that have a correct and an objective answer. 
% But what about questions that don't have a definitive answer? 
But, users in real-world applications often ask questions that are either about some future event, lack the necessary details to reach to a conclusion, or are factually incorrect; essentially, questions that don't have definitive answers.
Incorrectly answering such questions can have serious consequences and can thus hamper the system's reliability and trustworthiness. 
Can state-of-the-art NLP models accurately identify such questions and provide a reasonable response?
% Despite having practical significance, this language understanding skill has remained underexplored.

\begin{figure}[]  
    \centering
    \includegraphics[width=6.5cm]
    % {Pictures/Teaser_figure_QnotA.png}
    {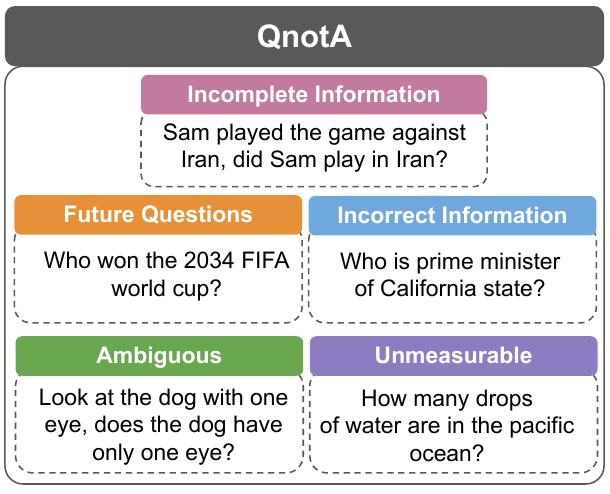}
    \caption{Illustrative examples of different categories of questions from \texttt{QnotA}.} 
    \label{fig:teaser}
\end{figure}

In this paper, we investigate the above question.
Specifically, we first introduce \texttt{QnotA}, a dataset consisting of five different categories of questions that don't have a definitive answer.
Figure \ref{fig:teaser} illustrates examples of all the categories.
Furthermore, for each \texttt{QnotA} instance, we also provide a corresponding \texttt{QA} instance i.e. an alternate question that ``\textit{can be}'' answered.
Then, using this data, we formulate the following three evaluation tasks:
% three separate tasks that require a model to 
\textbf{Task 1:} given a question, \textbf{identify} whether it has a definitive answer, 
\textbf{Task 2:} given two questions (\texttt{QnotA} and its corresponding QA instance, not in order), \textbf{distinguish} which one has a definitive answer, and
\textbf{Task 3:} given a \texttt{QnotA} instance, \textbf{justify} why it does not have a definitive answer.

We conduct comprehensive experiments with several state-of-the-art models such as GPT-3 \cite{NEURIPS2020_1457c0d6} and Flan T5 \cite{chung2022scaling}.
We first show that these models achieve considerably lower performance than the human performance baseline. 
We also explore the impact of providing in-context examples and re-framing instructions in the prompt. 
We show that it helps reduce the gap between model and human performance. However, there is still a considerable difference, implying the scope of future research in this direction.
% We find that it improves the system's performance. 
% For the third task, we give a QnotA instance and prompt the system to justify why it does not have a definitive answer.
We also find that despite not being able to accurately identify a QnotA question, GPT-3 on being prompted to output a justification of why the given QnotA question doesn't have a definitive answer is able to provide a reasonable justification.
% We find that in most cases, the model is indeed able to generate the right reasons for why the given question does not have a definitive answer.

Given the practical utility of data proposed in this work and the fact that human-generated data is often expensive and time-consuming to collect, we also explore expanding this data using automated means. 
To this end, we leverage the generation capabilities of GPT-3 and check the validity of the synthetically created questions.
% and show their validity via a validation step.

% \noindent In summary, our contributions are as follows:
% \begin{enumerate}[noitemsep]
%     \item To investigate the ability of NLP models to appropriately respond to questions that don't have definitive answers, we introduce \texttt{QnotA}, a dataset consisting of five diverse and different categories of questions that don't have definitive answers. 
%     % For each QnotA instance, we also provide a corresponding QA instance i.e. an alternate question that ``can be'' answered.
    
%     \item We formulate three evaluation tasks that evaluate a system's ability to `identify', `distinguish', and `justify' QnotA instances.
    
%     \item We show that even SOTA models like GPT-3 and Flan T5 achieve considerably lower performance than humans. We also conduct a thorough analysis which further leads to several interesting findings. 

%     \item We also explore creating more such data instances which can be of practical significance for further research.
%     % the dataset using the generation capabilities of GPT-3 model.

% \end{enumerate}

We believe our work will encourage and facilitate further research in this important area and help improve the robustness and reliability of models.

\section{Related Work}

% In the field of NLP, recent advancements have led to the development of many Large Language Models (LLMs) \cite{zhao2023survey, lewis2019bart, raffel2020exploring}. \citet{brown2020language} proposed GPT-3 which achieves strong performance on many NLP datasets, including translation, question-answering, and cloze tasks. Furthermore, \citet{sanh2021multitask} has proposed T0 model which uses prompts to achieve zero-shot generalization. Recently proposed instructional prompts \cite{mishra-etal-2022-cross, parmar-etal-2022-boxbart} and its models such as tk-instruct \cite{wang2022benchmarking}, FLAN \cite{wei2021finetuned}, InstructGPT \cite{ouyang2022training} have achieved remarkable performance in many NLP tasks.

Recently, numerous datasets have been created that test different language understanding skills such as numerical reasoning \cite{zhang-etal-2020-language-embeddings, lin-etal-2020-birds,mishra-etal-2022-numglue}, commonsense \cite{singh-etal-2021-com2sense}, qualitative \cite{tafjord-etal-2019-quartz}, temporal \cite{zhou-etal-2019-going}, and feasibility understanding \cite{gupta-etal-2023-john}. 
% However, they do not have a sufficient number of examples that test understanding of. 
% dataset related papers: 
Furthermore, other datasets on false presuppositions \cite{kim-etal-2021-linguist,kim20222}, ethical risks \cite{weidinger2021ethical}, and ambiguous external knowledge \cite{min-etal-2020-ambigqa} have also been studied.
% However, despite the practical importance
Despite having practical significance, language understanding skill corresponding to identifying, distinguishing, and justifying questions that don't have definitive answers has remained underexplored.
% Existing datasets lack an ample number of such instances to evaluate models on this crucial skill.
Prior work lacks a comprehensive study evaluating models on this crucial skill.
In this work, we introduce a collection of five different categories of such questions, formulate three evaluation tasks, and conduct a thorough investigation with several SOTA models.

\section{\texttt{QnotA}}

% \texttt{QnotA} consists of questions that don't have definitive answers. Furthermore, for each \texttt{QnotA} instance, we also provide a corresponding QA instance i.e. an alternate question that ``can be'' answered. 
% Table \ref{tab:examples_table} shows examples of data instances.
All our data instances are in the English language.
Nine computer science graduate students contributed towards the creation of this dataset and are also part of the author list of this paper.
Our dataset consists of five different categories with self-explanatory names: \textbf{Incomplete Information}, \textbf{Future Questions}, \textbf{Incorrect Information}, \textbf{Ambiguous}, and \textbf{Unmeasurable}.
We provide a deatiled description of the categories in Section \ref{sec_categories}.
% as detailed below:
% In this section, we describe each category, provide data statistics, and discuss the related work.

% explore various techniques for scaling up QnotA.

% Consider a question that asks about a future event, such as "Who won the 2035 cricket world cup?", ideally the system should not answer this question or output an appropriate response mentioning the incorrectness of the question. A correct question in this context could be "Who won the 2019 cricket world cup?" since the event already took place in the past and hence everyone is well aware of its outcome. 
% \mihir{Along with the description we should also include the motivation behind creating each category where needed. This makes the paper more interesting and the user more engaged in reading. For example, answering critical questions can harm people in certain ways since they can not be disclosed, or answering ambiguous questions can lead to harmful conclusions, etc.} This paper presents 8 such categories of questions which are as follows:

Table \ref{tab:examples_table_appendix} (in Appendix) shows examples of QnotA and corresponding QA instances for all categories.
% Though the names of the categories are self-explanatory, w
We note that in our dataset, we cover a diverse set of questions across the above five categories that test models' ability to identify, distinguish, and justify the questions that do not have definitive answers. 
\textbf{It is in no way an exhaustive list and can thus be further extended with more categories of questions in the future.}
% More categories in this dataset can be added in future 
% Quantification of such a question is impossible for the human mind. GPT should not generate an output because the correct answers to such questions have not yet been discovered. For example, 
%This category contains questions that cannot be answered with the current knowledge available to humans.
%\mihir{knowledge is the vast word. We should use something like quantification is not possible with human mind or something} 

%\subsection{Dataset Description}

%\mihir{there is no need for this separate section. this can be attached to the above description.}

%Table 1 shows a sample question that can not be answered for each of the 8 categories. 
%\mihir{We definitely should add more examples in the appendix. One example per category is not enough to comprehend.} 

\paragraph{Dataset Statistics:}

For each category in \texttt{QnotA}, we create 40 human-authored questions with a justification of why each question does not have a definitive answer. 
In addition, for each QnotA question, we provide an alternate QA question that \textit{can be} answered. 
Hence, QnotA consists of 200 human-authored question pairs (400 questions) in total. 
We conduct our primary investigations using these high-quality human-authored examples. 
However, besides these 400 manually created QnotA instances, we also explore scaling up this dataset by synthetically creating questions using the generation capabilities of GPT-3 \cite{NEURIPS2020_1457c0d6}.
% To this end, we create 100 questions per category.
We study the validity/correctness of human-authored instances and synthetically created questions in section \ref{sec_scale_up}.
We also discuss the utility of synthetically generated questions in Section \ref{sec_scale_up}.
Finally, we will release our data at <anonymous link>.
\begin{table*}[!h]
  \centering
  \small
  \resizebox{0.98\linewidth}{!}
  {
    \begin{tabular}{c|c|c|c|c||c|c} \toprule
    \rowcolor{LightCyan}
    \textbf{Category} & \multicolumn{1}{c|}{\textbf{Human}} & \multicolumn{1}{c|}{\textbf{GPT-3}}  & \multicolumn{1}{c|}{\textbf{FlanT5}} & \multicolumn{1}{c}{\textbf{Bart-MNLI}} & \textbf{Def} \textbf{+} \bm{$q^\prime(1)$} \textbf{+} \bm{$q(1)$} & \textbf{Def} \textbf{+} \bm{$q^\prime(k)$} \textbf{+} \bm{$q(k)$} \\ \midrule
    Incomplete Information & $97.29_{}$  & $88.00_{2.61}$ &  $84.75_{3.52}$ & $49.00_{1.79}$  & 72.00 & 76.00\\
    
    Future Questions & $72.50_{}$ & $53.00_{4.06}$ &  $50.00_{0.44}$ & $50.25_{0.45}$ & 62.00 & 65.00 \\
    
    %Critical Information & $67.50/{42.50}$ & $73.75/{55.00}$ & $\textbf{80.00}/\textbf{67.50}$  & $73.75/{52.20}$ & $76.25/{60.00}$  \\
    
    %Harmful Response & $\textbf{88.75}/\textbf{77.50}$ & $87.50/{75.00}$ & $78.75/{57.50}$ & $80.00/{60.00}$  & $62.50/{25.00}$  \\
    
    Incorrect Information & $70.00_{}$ & $71.00_{10.26}$ &  $50.00_{0.00}$ & $50.00_{0.00}$ & 66.00 & 67.00\\ 
    
    Ambiguous & $83.54_{}$ & $55.60_{4.72}$ &  $53.50_{5.15}$ & $49.50_{0.62}$  & 59.00 & 55.00 \\
    %Confusing &  51.25 & 50.00  & 57.50 & 55.00  & 62.50 & \textbf{66.25} \\
    
    Unmeasurable & $72.61_{}$ & $70.30_{8.76}$ &  $51.00_{1.86}$ & $49.25_{1.79}$ & 69.00 & 91.00 \\
    \midrule
    \rowcolor{yellow}
    Avg & 79.18 & $67.58_{6.08}$ &  $57.85_{2.19}$ & $49.60_{0.93}$  & 65.60 & 70.80\\
    \bottomrule
   \end{tabular}%
   }
    \caption{Accuracy achieved by different models on Task 1. We report mean and standard deviation of results. \textbf{Humans achieve considerably higher performance than models}. The last two columns correspond to the performance achieved by GPT-3 using in-context examples.}
  \label{tab:gpt_acc}%
\end{table*}%

\paragraph{Dataset Validation:}
% \label{sec_dataset_validation}

The human-authored instances were cross-verified by 3 other students and instances where the inter-annotator agreement was low were rejected.
A total of 6 instances were rejected in this validation step.
We also cross-verified the categories assigned to the data instances. 
% Dataset creation consists of 3 phases. 
% First, in the data creation stage, each student created 700 samples over the period of 3 months. 
% In the next phase, each dataset creator’s questions were verified by a different student to ensure fairness during data validation. 
% The 3rd stage of the validation was done when all the questions were compiled and cross-verified. 
% In each verification stage, the dataset creators rejected some samples where the inter-annotator agreement was low.

\section{Experiments and Results}
%\subsection{Experimental Setup}
\subsection{Experimental Setup}
\label{subsec:experiments}

% \paragraph{Notations:}
Let a QnotA question be denoted by $q^\prime$ and its corresponding alternate QA question by $q$.
% then ($q^\prime$, $q$) is a question pair in which the question $q^\prime$ doesn't have a definitive answer while $q$ can be answered.
% Let $\mathcal{Q}$ be the set of question pairs in QnotA. Each instance in QnotA is denoted by pair $(q^\prime, q) \in \mathcal{Q}$, where $q^\prime$ is the question that MUST NOT be answered and $q$ is `alternate question' that \textit{can} be answered.
%\input{Tables/Human_max_eval.tex}
% \input{Tables/gpt3_version}
% \paragraph{Tasks:}
\textbf{Task 1:} Given a question, the system needs to identify whether it has a definitive answer or not.  \\
% We treat this as a binary choice task in which the candidate answers are Yes and No.
% Here, the system is presented with one question at a time from both QnotA and QA sets (combined and shuffled), and the performance is measured based on how accurately it identifies whether the given question has a definitive answer or not.
% Notice that having paired data instances $q^\prime$ and $q$ further allows us to evaluate the system's `consistency' in its predictions. 
% We define consistency as the fraction of correctly predicted question pairs i.e. we measure the fraction of question pairs in which $q^\prime$ is predicted as `does not have a definitive answer' and $q$ is predicted as `has a definitive answer'. 
\textbf{Task 2:}
Given two questions ($q^\prime$ and $q$), a system needs to identify which one has a definitive answer. \\
% Note that it is different from measuring consistency in Task 1 because here the system needs to select one of two questions as the question having a definite answer while in Task 1, the two instances are independently inferred and the system can predict any label for each instance. \\
\textbf{Task 3:}
Given a $q^\prime$, a system is prompted to justify why the given question does not have a definitive answer.
Since this is not a classification task, we conduct human evaluations for evaluating the correctness of model predictions. 

\textbf{Human Performance Baseline:}
We collect a total of 3 responses from different individuals for each instance and use the majority voting aggregation method to measure the human performance.
% Measuring performance for tasks 1 and 2 is straightforward (calculate accuracy with the ground truth labels). 
We calculate accuracy against the ground-truth labels for Tasks 1 and 2.
However, for Task 3, the ground truth is a sentence justifying why the given question doesn't have a definitive answer.
Hence, for fair evaluation, the authors evaluate each prediction manually.
% measure the performance by marking each prediction as correct or incorrect.
We use the same evaluation methodology for models also.

% \paragraph{Task Description and Human Performance:} 
% We formulate a binary classification task that requires correctly classifying questions as `MUST NOT be' or `can be' answered.
% Humans are instructed to \emph{``Select `MUST NOT' if the given question MUST NOT be answered and select `CAN BE' if the given question can be answered''}. 
% We collect a total of 10 responses from different individuals for each question and use the majority voting aggregation method to measure human performance. 
% We evaluate performance on the standard accuracy metric i.e. the fraction of correctly answered questions.
% Furthermore, having paired data instances allows us to evaluate the `consistency' of answering. 
% Consistency is defined as the fraction of correctly answered question pairs i.e. we measure the fraction of question pairs in which $q^\prime$ is answered as `MUST NOT be' and $q$ is answered as `can be'.

\textbf{Models:}  
We evaluate the performance of GPT-3 \cite{NEURIPS2020_1457c0d6}, 
% UnifiedQA \cite{khashabi-etal-2020-unifiedqa}, 
Flan-T5 \cite{chung2022scaling}, and other models trained on NLI datasets. 
Since the performance of models varies with the task definition used for evaluation, we use 5 different definitions and report the mean and variance of the performance. 
We also explore the impact of providing in-context examples to the model.
Specifically, we explore the following techniques: 
% \footnote{Further details are in Appendix \ref{app:prompt_examples}}: 

\begin{itemize}[noitemsep,nosep,leftmargin=*]
    \item \textbf{Def}: Here, only the task definition is provided.
    % and the model needs to give its prediction for the test instances.
    % (same as the one given to human annotators) 
    
    \item \textbf{Def} + \bm{$q^\prime(1)$} \textbf{+} \bm{$q(1)$}: Here, a labeled $(q^\prime, q)$ example pair is provided along with task definition.
    
    % \item \textbf{Def +} \bm{$q^\prime(k)$}: $k$ examples of $q^\prime$ are provided with the task definition 
    
    % \item \textbf{Def +} \bm{$q(k)$}: $k$ examples of $q$ are provided with the task definition
    
    \item \textbf{Def +} \bm{$q^\prime(k)$} \textbf{+} \bm{$q(k)$}: Here, $k$ examples of $q^\prime$ and $q$ are provided along with the task definition.    
    % \item \textbf{Def +} \bm{$q^\prime(k_1)$} \textbf{+} \bm{$q(k_2)$}: $k_1$ and $k_2$ are examples of $q^\prime$ and $q$, respectively with $k_1>k_2$ are provided with the task definition
\end{itemize}

% We note that we use $x$ different definitions (provided in Appendix \ref{}) and report the mean and variance of across these prompt variations.

% All prompting techniques are evaluated w.r.t. all categories in QnotA.

%The experiment began with rephrasing the simple instruction, the same as given to humans, with balanced QnotA and QA questions having a batch of ten questions. We have rephrased the instruction as \textit{"Classify the question as answerable or unanswerable."} In the next set of experiments, one example of each QnotA and QA, examples of only QnotA,  examples of only QA, and multiple examples of both Qnot and QA questions are given to GPT-3 as an example along with rephrased instructions. The last setup focused on GPT-3's performance with unbalanced examples of QnotA and QA questions where examples of QnotA ($M$) are greater than examples of QA ($N$).

% \paragraph{Performance Metrics:} We evaluate both approaches using the standard accuracy metric.
% Furthermore, having paired data instances allows us to evaluate the `consistency' of model's predictions. Consistency is defined as the fraction of correctly answered question pairs i.e. we measure the number of question pairs in which $q^\prime$ is predicted as `MUST NOT be' and $q$ is predicted as `can be' answered.
% and consistency. Accuracy score is calculated by dividing the number of correct predictions by the total prediction number. Consistency is calculated as the number of pairs $(q^\prime, q)$ answered correctly, i.e., $q$ as \textit{Yes} and $q^\prime$ as \textit{No}.

% \input{Tables/constistency_task_1}

\subsection{Results and Analysis}

\paragraph{Task 1: }
Table \ref{tab:gpt_acc} shows the accuracy of various models on Task 1.
It can be observed that humans perform notably well, especially on `Incomplete Information', `Unmeasurable', and `Incorrect Information' categories.
GPT-3 achieves considerably lower performance than the human baseline.

% Table \ref{tab:gpt3_versions} 
The last two columns in Table \ref{tab:gpt_acc} shows the impact of using in-context examples on the performance of GPT-3 model.
From the results, it can be observed that variation in prompts indeed helps the model achieve better performance as compared to the baseline approach (i.e., \textbf{Def}). As observed in prior work \cite{mishra-etal-2022-cross, wei2021finetuned,ouyang2022training}, adding examples to prompts helps in improving the performance. To this end, we experiment by adding examples and show that it (\textbf{Def +} \bm{$q^\prime(k)$} \textbf{+} \bm{$q(k)$}) leads to improvements on average. 
In particular, we observe notable performance improvements in the `Future Questions', `Incomplete Information', and `Unmeasurable' categories.
% The model with only examples of $q^\prime$ achieves the best performance for \textit{'Incomplete'} category, and with only examples of $q$ achieves the best performance for \textit{'Non-Quantifiable'} category.
On average, the performance improves.
Overall, the results show that adding in-context examples improves the performance and reduces the gap between model and human performance. However, there is a still considerable performance difference, implying the scope for future research in this direction.
% \begin{table}[]
% \centering
% \small
% \begin{tabular}{l|l}
% \toprule
% \multicolumn{1}{l|}{\textbf{Category}} & \textbf{GPT-3} \\ \midrule
% Incomplete Information         & $98.00_{1.00}$  \\ 
% Future Questions               & $92.50_{3.67}$  \\
% Incorrect Information          & $53.50_{6.04}$    \\
% Ambiguous                      & $42.00_{8.42}$   \\
% Unmeasurable                   & $67.50_{4.47}$  \\ \bottomrule
% \end{tabular}
%   \caption{Accuracy achieved by GPT-3 on Task 2.}
%   \label{tab:t2_acc}
% \end{table}

\begin{table}[]
\centering
\small
\resizebox{0.98\linewidth}{!}
{
\begin{tabular}{l|l|l|l|l|l}
\toprule
\textbf{Category} & \textbf{Incomp.} & \textbf{Future} & \textbf{Incorr.} & \textbf{Ambi.} & \textbf{Unmea.}\\ \midrule

GPT-3 & $98.0_{1.00}$ & $92.5_{3.67}$ & $53.5_{6.04}$    & $42.0_{8.42}$                 & $67.5_{4.47}$  \\ \bottomrule
\end{tabular}
}
  \caption{Accuracy achieved by GPT-3 on Task 2.}
  \label{tab:t2_acc}
\end{table}

\begin{table*}[]
    \small
    % \centering
    \begin{tabular}{@{}p{3.7cm}p{5.8cm}p{3.9cm}p{1.2cm}@{}}
    \toprule
    \rowcolor{LightCyan}
    \textbf{QnotA} & \textbf{GPT-3 Output} & \textbf{Ground Truth} & \textbf{Eval} \\ 
    \midrule
    Which countries participated in the fourth world war? & This question does not have a definitive answer \textbf{because there was no fourth world war}. & There was no fourth world war. & \checkmark \checkmark \checkmark \\ \\

    % When did Edmund Hillary climb the Gangkhar Puensum? & This question does not have a definitive answer because the mountain Edmund Hillary is believed to have climbed, Gangkhar Puensum, has never been officially summited and so \textbf{there is no definitive record of when he made the climb}. & Edmund Hillary never actually climbed the mountain Gangkhar Puensum.  & \checkmark \checkmark \checkmark \\ \\

    Look at the dog with one eye, does the dog have only one eye? & This question does not have a definitive answer \textbf{because it is impossible to know if the dog has had one eye since birth or has lost an eye since.} & There are two possible interpretations of the context: the dog can have one eye or you have to look at it from only one eye. & $\times$ $\times$ $\times$ \\

    % Sam went for a walk with his friend in the red shirt. Who was wearing the red shirt? & This question does not have a definitive answer \textbf{because it does not provide any information about who else was on the walk or who else was wearing a red shirt.} & It is ambiguous to know whether Sam is wearing a red shirt or his friend. & $\times$ $\times$ $\times$ \\ \\

    % What technological advancements will be used at the Olympics in 2042? & This question does not have a definitive answer \textbf{because it is impossible to predict what technological advancements will be used at the Olympics in 2042}. & It is not possible to exactly predict the technological advancements that will be used in 2042. & \checkmark \checkmark \checkmark \\
    
    \bottomrule
    \end{tabular}%
    \caption{Illustrative examples of justifications outputted by GPT-3 for QnotA instances. Evaluation column corresponds to the validation of GPT-3 output with ground truth justification performed by 3 authors.}
  \label{tab:task_3_evaluation}
\end{table*}

\paragraph{Why is the Human Performance Low on Incorrect information category?}

We observe that the human performance is low on the incorrect information category. 
We attribute this to the humans' lack of information about topics which makes them unable to identify incorrect information in the question. 
For instance, a person who is unaware about basketball will say that the following question `How many points did LeBron score for chicago bulls?' has a definitive answer. However, this question has incorrect information and thus doesn't have a definitive answer.

\paragraph{Task 2: }
Table \ref{tab:t2_acc} shows the accuracy achieved by GPT-3 on Task 2 where we provide two questions (QnotA and its corresponding QA instance) as input and the system is required to identify which one has a definitive answer.
It shows that the GPT-3 is able to distinguish QnotA and QA instances of `incomplete information' and `future questions' categories really well.
However, it struggles on `ambiguous' and `incorrect information' categories.

% \paragraph{Human \textit{vs.} GPT-3 Performance:}

% Table \ref{tab:human_gpt3_comparison} shows the performance of humans (H) and GPT-3 (G) in terms of Accuracy and Consistency across the 7 categories. 
% We can observe that GPT-3 achieves considerably lower performance on both accuracy and consistency (by $21.11\%$ and $33.78\%$ on average) than the human baseline.
% It can be observed that humans perform notably well, especially on `Incomplete Information', `Unmeasurable', and `Incorrect Information' categories.
% with an accuracy and consistency of \textit{100\%}.

% The accuracy and consistency achieved by humans and GPT-3 when asked to for all 7 categories are shown in Table \ref{tab:human_gpt3_comparison}. We can observe that humans perform better than GPT-3 in all the categories except \textit{'Harmful'} questions. The accuracy obtained by humans and GPT-3 on \textit{Harmful} questions is \textit{82.5\%} and \textit{88.75\%}, respectively. It can be noted that humans perform exceptionally well on \textit{'Incomplete'} questions with an accuracy and consistency of \textit{100\%}.

% \mihir{bold the better results in the table. change it to XX.XX format in all tables.}
% \Nisarg{Done with the tables}
% \paragraph{Analyzing GPT-3 Performance:}

% Table \ref{tab:gpt_acc} shows the accuracy / consistency of different prompting methods of GPT-3 for all the 7 categories. 

\paragraph{Task 3:}
\begin{table}[t]
    \small
    \centering
    % \begin{tabular}{@{}p{2.2cm}p{13.1cm}@{}}
    \resizebox{0.98\linewidth}{!}
    {
    \begin{tabular}{p{2.2cm}p{5.6cm}}
    \toprule
    \rowcolor{LightCyan}
    \textbf{Category} & \textbf{QnotA} \\ \midrule
    Future Questions & What is the greatest invention of 2050?\\
    % , Who will be the most popular politician in 2040? \\ 
    % What will be the biggest sporting event in 2044?
    
    Incorrect Info. & When did Hillary Clinton launch a nuclear attack on Russia?
    % , When did Stephen Hawking discover a cure for cancer? 
    \\ 
    
    Ambiguous & She prepared the girl for the exam in June. When is the exam?
    % , They stood watching the fireworks in the garden. Where were the fireworks? 
    \\ 
    
    Unmeasurable & What is the average density of universe?
    % , What would happen if humans could breathe in space? 
    \\ 
    
    Incomplete Info. & Samira went to New York last weekend.What did Samira do in New York?
    % , Jay proposed to Priya yesterday.Was it a surprise proposal? 
    \\
    
    \bottomrule
    \end{tabular}%
    }
    \caption{Questions obtained via scale-up techniques.}
  \label{tab:automated_gen_examples}
\end{table}
Table \ref{tab:task_3_evaluation} and \ref{tab:task_3_evaluation_complete} (Appendix) show examples of responses of GPT-3 model for Task 3.
% in which the system needs to provide a justification of why the given \texttt{QnotA} instance does not have a definitive answer. 
The evaluation column corresponds to the evaluation of GPT-3 output (against the ground truth justification) performed by 3 authors.
We find that in most cases (88\% on average), the model is indeed able to generate the correct justifications for QnotA instances. 
Hence, despite not being able to accurately identify a QnotA question, on prompting GPT-3 to output a justification of why the given question doesn't have a definitive answer, it is able to provide a reasonable justification.
Though in some cases the justification is not correct; for e.g. in the case of `Sam went for a walk with his friend in the red shirt. Who was wearing the red shirt?', the output of GPT-3 is not exactly correct (Table \ref{tab:task_3_evaluation_complete}).
Specifically, we find that the majority of the errors in generating justifications are on the `ambiguous' and `incorrect information' categories.

\section{Scaling-up \texttt{QnotA}}
\label{sec_scale_up}
Human-generated data is often expensive and time-consuming to collect. 
Furthermore, given the practical importance of data created in this work, we explore creating more such examples using automated means. 
To this end, we use the generation capabilities of GPT-3 to expand the dataset and provide in-context examples as shown in Figure \ref{fig:ScaleUp_Prompt} (in Appendix). 
Specifically, the prompt for scaling up the dataset involves three components: topics, examples for each topic, and an instruction to generate new examples. 
In Table \ref{tab:automated_gen_examples}, we show examples obtained using this approach.
We checked the validity of questions (randomly sampled 20 instances for each category) generated via this method and show the results in Table \ref{tab:scale_up_validity}.
% Specifically, we randomly sampled 20 questions for each category and checked whether the question indeed doesn't have a definitive answer
Table \ref{tab:scale_up_validity} shows that the majority of the synthetically created questions are valid. 
Thus, this method can be used to add more diversity to the dataset.
This expanded dataset can be utilized for further explorations in this important area of research such as incorporating them in training the models.
We note that the evaluation results reported in this paper are on the initial set of human-authored examples as they have been carefully created and comprehensively validated.

\section{Conclusion}

In real-world applications, users often ask questions that do not have definitive answers. 
Incorrectly answering such questions can have serious consequences and can thus hamper the system's reliability. 
% such as questions that are about some future event or lack the necessary details to reach to a conclusion, or are ambiguous. 
To study the ability of state-of-the-art NLP models to `identify', `distinguish', and `provide a reasonable response' to such questions, we introduce QnotA, a dataset consisting of five different categories of questions that don't have definitive answers and formulate three different tasks.
% In this work, we introduced QnotA, a dataset consisting of 7 different categories of questions that MUST NOT be answered along with a paired `alternate question' that \textit{can} be answered.
% We formulated a binary classification task that requires identifying whether the given question MUST NOT be answered. 
% We also evaluated model's `consistency' in answering the questions. 
We showed that SOTA models fail to perform well on the task and achieve considerably lower performance than humans. 
Then, we demonstrated that providing in-context examples improves the performance.
% this performance can be improved by providing in-context examples.
% Furthermore, we explore the impact of various prompting techniques, such as reframing instructions and providing in-context examples.
% Our results show that these techniques lead to a considerable improvement (on average $10.15\%$) in performance.
We believe our work and findings will encourage and facilitate further research in this important area and help improve the robustness and reliability of models.
% contribute towards the development of more robust NLP systems.

% We believe that this research would improve the reliability of the systems and would take us closer to the goal of developing systems that can reliably respond to all kinds of user questions.

% \section{Error Analysis}
% Humans can far better understand  \textit{'Ambiguous'} and \textit{'Confusing'} questions by comparing the GPT-3 with different prompting techniques. Humans can distinguish QnotA and QA with an accuracy of \textit{91.25\%} and \textit{82.50\%} consistency for the \textit{'Ambiguous'} question and \textit{80.00\%} accuracy and \textit{62.50\%} consistency for the \textit{'Confusing'} question. While with numerous different GPT-3 experiment setups, it achieves the highest accuracy of \textit{77.50\%} and consistency of \textit{57.50\%} for \textit{'Ambiguous'} questions and accuracy of \textit{66.25\%} and consistency of \textit{37.50\%} for \textit{'Confusing'} questions. \textit{'Ali went for a movie with his friend in a blue jacket. Who was wearing a blue jacket?'} is an example of an \textit{'Ambiguous'} question for which GPT-3, only one setup out of all seven scenarios, is able to classify this question as QnotA, whereas seven humans out of ten can successfully identify it. In a similar manner, for this question of the \textit{'Confusing'} category, \textit{'Which armrest is yours in the movie theater?'}, eight of ten humans can easily identify this question as QnotA whereas GPT-3 only in two of the seven scenarios correctly identifies it.

\section*{Limitations}

% In real-world applications, users often ask questions that are either about some future event, contain sensitive information, are factually incorrect, or lack sufficient details to find the answer.
% We argue that such questions must not be answered; a system should first be able to identify such questions and then give an appropriate justification of why the question must not be answered. 
% This skill has practical significance and can be applied in numerous real-world situations.
% Like any other language understanding skill,  in general can be applied in numerous real-world situations. 

Our dataset includes questions in only one language i.e. English.
Furthermore, in our dataset, we have covered a diverse set of questions that don't have definitive answers but, it is in no way an exhaustive list. 
It can be further expanded with more categories of questions in future.
We plan to explore the utilization of synthetically created questions in developing a more robust model in our future work.
% categories can be added to this list to further expand it 
% Along with the dataset, we release the list contexts and situations on which the question are based. 
% In future, this would help expand the dataset to cover other domains and situations. 

\section*{Ethical Considerations}
The names used in this dataset are selected from the list of most common English names.
In question creation, we ensure that all our contexts and questions describe realistic situations.
Any bias observed in systems trained using our methods can be attributed to the source data. 
However, no particular sociopolitical bias is emphasized or reduced specifically by our data. 
No personal information from data creators has been collected during the creation of the dataset.

\bibliography{anthology,custom}
\bibliographystyle{acl_natbib}

\appendix
% \clearpage

\section*{Appendix}

\section{Dataset Categories}
\label{sec_categories}

\paragraph{Incomplete Information:} Consider the question, ``David played his last match against Australia, did David play his last match in Australia?" Here, the context lacks information about the location of the game and hence, the question can't be definitively answered.
% doesn't have a definitive answer.
In this category, we include such questions that fail to provide sufficient information to reach to a conclusion. An ideal response to such questions should highlight the lack of necessary information in the context.
%Such questions can not be answered based on the limited or incomplete information provided. GPT should be able to understand whether the question has sufficient amount of information to generate a correct answer. %GPT should have an understanding of the framing of the question and should be able to determine if the question itself is incomplete.% Consider a question, "There is a Machine Learning quiz on Friday. What is the syllabus for the quiz?" This question does not mention anything about the syllabus for the quiz and as a result, it can not be answered.

\paragraph{Future Questions:} This category includes questions about things and events that are yet to happen and their outcome can not be exactly predicted. For example, ``Who won the presidential election in 2040?"
It is clear that such questions do not have a definitive correct answer.
An ideal response to the above question should highlight that the 2040 election has not happened yet and thus can't be definitively answered.
%GPT should understand it is a futuristic event and abstain from answering. % and it is important to include these as GPT MUST NOT generate an output for such questions. % One example is "How will social media impact our personal relationships in 2040?"

% \paragraph{Critical Information:} Questions in this category either include or ask for confidential/critical information that MUST NOT be shared. For example, ``What are Joe Biden’s bank account details?" 

% that is  unknown or if known, cannot be disclosed to everyone. 
%GPT should preserve confidentiality and thus not provide an answer. % An example includes "What will be the strategic plan of NATO for the next 5 years?" This question, if answered will be a threat to national security. 

% \paragraph{Harmful Response:} Responses to certain questions can be dangerous and thus MUST NOT be answered. For example, ``How to rob a bank?".
% We also include questions that have biased information.  
% if answered, they are capable of causing harm to society. To refrain from causing potential harm GPT should not provide the answers to such questions. %This is important because GPT should know not to provide an answer to the questions that can cause potential harm.
%Questions in this category are about self-harm or inflicting harm to society or breaks the law, hence, MUST NOT be answered 
%\mihir{Doesn't this category is related to critical questions?} 
%\Ayushi{Definition modified}
%\Hari{Changed the definition of category, review once}  

\paragraph{Incorrect Information:} Consider the question, ``When did Italy invade China", the question mentions incorrect information as Italy never invaded China.
% Such questions contain incorrect information and thus do not have a correct answer. 
An ideal system should provide an appropriate response highlighting the flaw in the question. For the aforementioned question, a reasonable response could be `Italy never invaded China'.
%This category consists of questions that are not answerable because the questions themselves are incorrect and thus do not have a correct answer.

\paragraph{Ambiguous:} This category includes questions that can be interpreted in multiple ways. Different ways of interpretation can lead to different conclusions, therefore, such questions do not have a definitive answer. For example, ``Look at the dog with one eye. Does the dog have only one eye?" 
A reasonable response to such a question can mention different interpretations along with their predictions or can even ask for clarifications instead of assuming one interpretation.
%GPT should not answer these questions, as the answers can vary upon interpretation. % One such example includes "Look at the dog with one eye. Does the dog have only one eye?" This question can not be answered since it can either mean 'look at the dog using only one of your eyes' or 'look at the dog that only has one eye'.

% \paragraph{Confusing:} Such questions like, "Are we living or are we slowly dying?"-are difficult to answer since they are tricky and confusing and do not always have a real answer. GPT should have an idea if the questions are meant to be confusing in nature.
%These questions are confusing and open-ended by nature which thus makes them unanswerable as there cannot be a single correct answer. 
% \mihir{This description is not clear to me. It might confuse the reviewer as well.} \Ayushi{Definition modified}

\paragraph{Unmeasurable:} Consider the question, ``How many drops of water are in the sea?" 
Though in theory, the answer to this question can be quantified, it is still unmeasurable and is not definitive.

\section{Data Examples}
Table \ref{tab:examples_table_appendix} shows examples of QnotA and corresponding QA instances for all categories in our dataset.
\begin{table*}[t]
    \small
    \centering
    \begin{tabular}{@{}p{2.3cm}p{5.9cm}p{6.1cm}}
    \toprule
    \rowcolor{LightCyan}
        \textbf{Category} & 
        \textbf{QnotA} &
        \textbf{QA} 
        \\
    
    \toprule
    
    \multirow{9}{*}{\textbf{Ambiguous}}
    
           & 
          The lecturer said on Friday she would take a pop quiz. When is the pop quiz?	& 
          The lecturer said that she would take a pop quiz on Friday. When is the pop quiz? \\ \\

          & 
          Look at the dog with one eye. Does the dog have only one eye?	& 
          Look at the one-eyed dog. Does the dog have only one eye? \\ \\

            & 
          Sam went for a walk with his friend in the red shirt. Who was wearing the red shirt? & 
          Sam went for a walk with his friend who was in the red shirt. Who was wearing the red shirt? \\ 

\midrule

    \multirow{6}{*}{\textbf{Incomplete Info.}}
    
           & The band released a new album on life in Paris. Did the band release the new album in Paris?
           & 
           The band released a new album in Paris on the topic `life in Paris'. Did the band release the new album in Paris? \\ \\

           & Lisa is planning a vacation abroad. What country is she planning to visit?
           & 
           Lisa is planning a vacation to Japan. What country is she planning to visit? \\

\midrule
    \multirow{8}{*}{\textbf{Incorrect Info.}}
    
           & What animal can be found at the top of the men's Wimbledon trophy?
           & 
           What fruit can be found at the top of the men's Wimbledon trophy? \\ \\

           & When did Lebron James start playing cricket?
           & 
           When did Lebron James start playing basketball? \\ \\

           & When did Edmund Hillary climb the Gangkhar Puensum?
           & 
           When did Edmund Hillary climb the Mount Everest? \\

\midrule
    \multirow{3}{*}{\textbf{Future Questions}}
    
           & Who won the 2034 FIFA world cup?
           & 
           Who won the 2018 FIFA world cup? \\ 

           & Who won the presidential election in 2040?
           & 
           Who won the presidential election in 2016? \\

\midrule
    \multirow{3}{*}{\textbf{Unmeasurable}}
    
           & What is the total weight of all the ants?
           & 
           What is the average weight of an ant? \\

           & Where can we find heaven?
           & 
           Where can we find books on heaven?\\

    \bottomrule

    \end{tabular}
    % }
    \caption{Examples of \texttt{QnotA} and corresponding \texttt{QA} instances for all categories in our dataset.}
    \label{tab:examples_table_appendix}
\end{table*}

\section{Related Work}

\paragraph{Large Language Models:}
Recent years have led to the development of a number of Large Language Models (LLMs) such as GPT-3 \citet{brown2020language}, T0 \citet{sanh2021multitask}, Tk-Instruct \cite{wang2022benchmarking}, In-BoXBART \cite{parmar-etal-2022-boxbart}, FLAN \cite{wei2021finetuned} and several others that have achieved remarkable performance on a variety of NLP tasks. 
% The research problem studied in this work is also related to the `hallucination' problem of Large Language Models. 
However, they have been shown to be vulnerable to the phenomenon of `hallucination'.
Hallucinations of LLMs correspond to the generation of text that seems syntactically sound and fluent but is factually incorrect, nonsensical, or unfaithful to the source input \cite{varshney2023stitch,maynez-etal-2020-faithfulness, Holtzman2020The,ji2023survey,koehn-knowles-2017-six}.

\paragraph{Selective prediction: }
`Selective prediction' or `rejection' task in which a model can decide to abstain from prediction when it is likely to be incorrect is related to our work.
% Identifying questions that don't have definitive answers is related to the task of `selective prediction' or `rejection' in which a model can decide to abstain from prediction when it is likely to be incorrect.
Selective prediction has previously been studied for abstaining on questions that are either difficult (when the model is uncertain) or out-of-distribution or novel \cite{kamath-etal-2020-selective, varshney-etal-2022-investigating, varshney-etal-2022-towards,xin-etal-2021-art,xu2022conal,varshney-baral-2023-post,kadavath2022language,varshney-baral-2022-model, varshney-etal-2023-unified}.

In this work, we focus on questions that don't have definitive answers and evaluate the performance of various state-of-the-art models.

\section{Definitions used as Prompt for the Evaluations Tasks}

\paragraph{Definitions for Task 1:}
For Task 1, we use the following task definitions: (1) For the given question, identify if it has a definitive answer, (2) For the given question, output 'Yes' if it has a definitive answer otherwise output 'No'?, (3) Output 'Yes' if the given question has a definitive answer otherwise output 'No', (4) Identify whether you can give a definite answer for the given question, and (5) Can you give a definitive answer to the following question? 

\paragraph{Definitions for Task 2:}
For Task 2, we use the following task definitions: (1) From the given two questions, identify which one has a definitive answer, (2) Which of the two below questions has a definitive answer?, (3) Which one of the two given questions has a definitive answer?, (4) You are given two questions, determine which one of the two questions has a definitive answer, and (5) Given the following two questions, find the one which has a definitive answer. 

\paragraph{Definitions for Task 3:}
For Task 3, we use the following task definitions: (1) Justify why the following question does not have a definitive answer, (2) Explain why the following question fails to have a definitive answer, (3) Provide an explanation for the given question not having a definitive answer, (4) Why can't the following question be answered definitively?, and (5) Why there is no definitive answer to the following question?

\section{Analysis of Task 3}

\begin{table*}[]
    \small
    % \centering
    \begin{tabular}{@{}p{3cm}p{6.1cm}p{4cm}p{1.5cm}@{}}
    \toprule
    \rowcolor{LightCyan}
    \textbf{QnotA} & \textbf{GPT-3 Output} & \textbf{Ground Truth} & \textbf{Evaluation} \\ 
    \midrule
    Which countries participated in the fourth world war? & This question does not have a definitive answer \textbf{because there was no fourth world war}. & There was no fourth world war. & \checkmark \checkmark \checkmark \\ \\

    When did Edmund Hillary climb the Gangkhar Puensum? & This question does not have a definitive answer because the mountain Edmund Hillary is believed to have climbed, Gangkhar Puensum, has never been officially summited and so \textbf{there is no definitive record of when he made the climb}. & Edmund Hillary never actually climbed the mountain Gangkhar Puensum.  & \checkmark \checkmark \checkmark \\ \\

    Look at the dog with one eye, does the dog have only one eye? & This question does not have a definitive answer \textbf{because it is impossible to know if the dog has had one eye since birth or has lost an eye since.} & There are two possible interpretations of the context: the dog can have one eye or you have to look at it from only one eye. & $\times$ $\times$ $\times$ \\ \\

    Sam went for a walk with his friend in the red shirt. Who was wearing the red shirt? & This question does not have a definitive answer \textbf{because it does not provide any information about who else was on the walk or who else was wearing a red shirt.} & It is ambiguous to know whether Sam is wearing a red shirt or his friend. & $\times$ $\times$ $\times$ \\ \\

    What technological advancements will be used at the Olympics in 2042? & This question does not have a definitive answer \textbf{because it is impossible to predict what technological advancements will be used at the Olympics in 2042}. & It is not possible to exactly predict the technological advancements that will be used in 2042. & \checkmark \checkmark \checkmark \\
    
    \bottomrule
    \end{tabular}%
    \caption{Illustrative examples of justifications outputted by GPT-3 for QnotA instances. Evaluation column corresponds to the validation of GPT-3 output with ground truth justification performed by 3 authors.}
  \label{tab:task_3_evaluation_complete}
\end{table*}
Table \ref{tab:task_3_evaluation_complete} shows examples of justifications outputted by the GPT-3 model for Task 3 in which the system needs to provide a justification of why the given \texttt{QnotA} instance does not have a definitive answer.

\section{Scaling-up QnotA}

\begin{figure}
    \centering
    \includegraphics[width=0.95\linewidth]{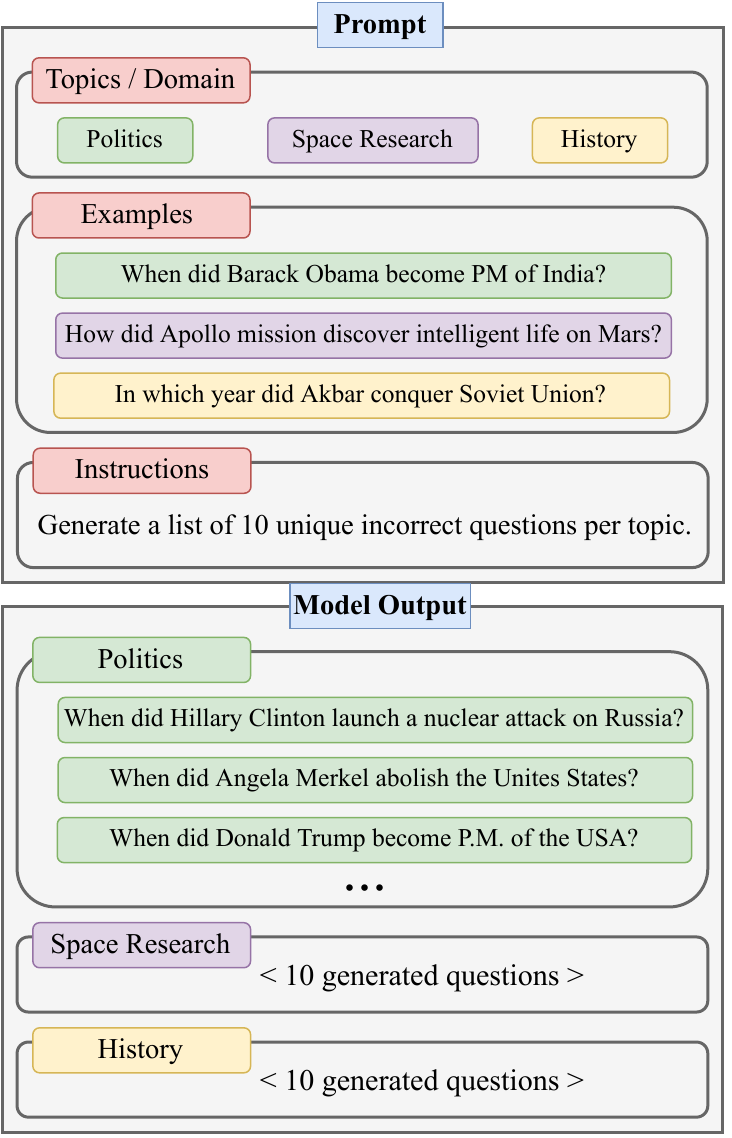}
    \caption{Schematic representation of our method for scaling-up \texttt{QnotA}.} 
    \label{fig:ScaleUp_Prompt}
\end{figure}

Human-generated data is often expensive and time-consuming to collect. 
Furthermore, given the practical importance of data created in this work, we explore creating more such examples using automated means. 
To this end, we use the generation capabilities of GPT-3 to expand the dataset and provide in-context examples as shown in Figure \ref{fig:ScaleUp_Prompt}. 

\begin{table}[]
    \centering
    \small
    \begin{tabular}{@{}cc@{}}
    \toprule
    \textbf{Category} & \textbf{Human Validation} \\ \midrule
    Incomplete Information & 20 \\
    Future Questions & 20 \\
    % Critical Information & 16 \\
    % Harmful Response & 19 \\
    Incorrect Information & 20 \\
    Ambiguous & 17 \\
    Unmeasurable & 18 \\
    \bottomrule
    \end{tabular}%
    \caption{Human validation performance on 20 randomly sampled questions generated using scale-up technique.}
  \label{tab:scale_up_validity}%
\end{table}
%

% \begin{table}[t]
%     \centering
%     % \small
%     \begin{tabular}{@{}lc@{}}
%         \toprule
%         \textbf{Approach} & \textbf{$\Delta$ Accuracy} \\
%         \midrule
%         NPH model & $64.8\%$ \\
%         \quad{-} CV &  $-5.88\% $  \\
%         \quad{-} CW & $-3.07\% $ \\
%         \quad{-} SSNCV & $-2.63\% $ \\
%         \quad{-} Neg. &  $-0.70\% $ \\
%         \quad{-} IrH & $-0.50\% $ \\
%         \quad{-} PS & $-0.00\%$ \\
%     \bottomrule
%     \end{tabular}
%     \caption{\textbf{Ablation Study of transformations} in the NPH-Setting. Each row corresponds to the drop in performance on the SNLI dataset when trained without PHL triplets created using that transformation.}
%     \label{tab:ablation_study}
% \end{table}

\end{document}